\pdfoutput=1
\documentclass[11pt]{article}
\usepackage{acl}
\usepackage[T1]{fontenc}
\usepackage[utf8]{inputenc}
\usepackage{times}
\usepackage{latexsym}
\usepackage{amsmath}
\usepackage{graphicx}
\usepackage{dsfont}
\usepackage{blindtext}
\usepackage{enumitem}

\usepackage{microtype}

\author{
    Dylan Ebert\\
    Brown University\\
    \texttt{dylan\_ebert@brown.edu}
    \\\And
    Chen Sun\\
    Brown University\\
    \texttt{chensun@brown.edu}
    \\\AND
    Ellie Pavlick\\
    Brown University\\
    \texttt{ellie\_pavlick@brown.edu}
}

\begin{document}

\title{Do Trajectories Encode Verb Meaning?}

\maketitle
\begin{abstract}
Distributional models learn representations of words from text, but are criticized for their lack of grounding, or the linking of text to the non-linguistic world. Grounded language models have had success in learning to connect concrete categories like nouns and adjectives to the world via images and videos, but can struggle to isolate the meaning of the verbs themselves from the context in which they typically occur. In this paper, we investigate the extent to which \textit{trajectories} (i.e. the position and rotation of objects over time) naturally encode verb semantics. We build a procedurally generated agent-object-interaction dataset, obtain human annotations for the verbs that occur in this data, and compare several methods for representation learning given the trajectories. We find that trajectories correlate as-is with some verbs (e.g., \textit{fall}), and that additional abstraction via self-supervised pretraining can further capture nuanced differences in verb meaning (e.g., \textit{roll} vs. \textit{slide}).
\end{abstract}

\section{Introduction}

While large distributional language models such as BERT \cite{devlin-etal-2019-bert} and GPT \cite{radford-2020-seeing,DBLP:journals/corr/abs-2005-14165} have had empirical success in deriving representations of words and sentences from large text corpora, most of these models lack \textit{grounding}, or a connection between the words and their real-world referents. Grounding, in addition to being necessary for multimodal tasks like video recognition, has been argued to lie at the core of language \textit{understanding} \cite{bender-koller-2020-climbing}. Work on grounded language learning associates language with the non-linguistic world, typically by learning from large-scale image \cite{bruni-etal-2011-distributional} or video \cite{sun2019videobert} datasets. 

Much prior work on language grounding has focused on concrete nouns (objects) and adjectives (attributes), which are captured well by patterns of pixels. Verbs, however, have received less attention, despite being essential for building models that can interact in realistic 3D environments \cite{shridhar-etal-2020-end,bisk-etal-2020-experience}. Verbs are especially challenging to model, given that they take place over time. Image and video data alone is insufficient to fully capture verb semantics, as demonstrated by prior work \cite{Yatskar_2016_CVPR}, in many cases failing to isolate the meaning of the verb from context in which it typically occurs. For example, \citealt{chao2018learning} show that an image of a person laying in the snow next to a snowboard is labeled ``standing on a snowboard". Moreover, recent work has introduced datasets and benchmarks based on situated 3D environments \cite{gan2020threedworld,deitke2020robothor,ebert-pavlick-2020-visuospatial,shridhar-etal-2020-end} that demonstrate the challenges of learning task-oriented behavior, which demands a combination of object and verb grounding.

In this paper, we test the hypothesis that the semantics of (concrete) verbs are grounded in the 3D trajectories of objects: i.e., the absolute and relative paths objects take through 3D space. We investigate if and when verb meanings appear to be a product of raw perception of objects in 3D space, and when differentiating verb meanings requires additional abstraction and representation beyond what is available via direct perception. To study this, we collect a clean dataset of 3D object trajectories in simulation. We collect human descriptions of these perceived world dynamics, i.e., to determine whether or not a given event constitutes a \textit{fall} or a \textit{tumble}. We then propose a self-supervised pretraining approach, whereby we train a time-series prediction model to obtain representations of trajectories in a 3D environment without any linguistic input. We evaluate the learned representations on how well they encode verb semantics for specific verbs. We show that the pretrained model learns to represent events in a way that aligns well with the meaning of English verbs, e.g. differentiating \textit{slide} from \textit{roll}. In summary, our primary contributions are:
\begin{enumerate}
    \item We introduce a new, clean dataset of 3D object trajectories paired with human judgments about whether or not each trajectory falls within the extension of each of 24 different verbs. To the best of our knowledge, this is the first dataset of its kind, and provides a valuable resource for empirical studies of lexical semantics. Our data is available at \url{https://github.com/dylanebert/simulated}.
    \item We compare several representation learning methods in terms of their ability to capture verb semantics \textit{without any linguistic signal during training}. In particular, we investigate the roll of abstraction (via self-supervised pretraining) compared to raw perception in capture verb meanings. To our knowledge, this is the first work to apply neural networks and (pre-linguistic) concept learning to the study of verb semantics.
\end{enumerate}

\begin{figure*}
    \centering
    \includegraphics{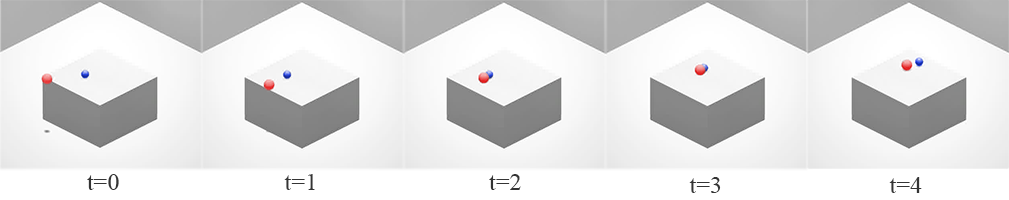}
    \caption{The \textit{Simulated Spatial Dataset} consists of procedurally generated motion data of a virtual agent interacting with an object. In this sequence the agent (red sphere) pushes the object (blue sphere). At t=0 and t=1, the agent approaches the ball. Then, in t=2 and t=3, the agent pushes to ball. Finally, at t=4, the ball is rolling away from the agent.}
    \label{fig:sequence}
\end{figure*}

\section{Related Work}

\paragraph{Grounded Language with Deep Learning.}
Our contributions add to a large body of work on grounded representation learning. Much of this work augments language modeling objectives with images \cite{silberer-lapata:2012:EMNLP-CoNLL,lazaridou2015combining,kiela2017learning} and videos \cite{sun2019videobert}. In this work, we focus on representations that encode verb semantics. Prior work on verb learning has been conducted in the computer vision community, typically described as ``human-object interactions" \cite{regneri2013grounding,chao2018learning,sun2018actorcentric,ji2019action}. Most closely related to our approach, which focuses on trajectory data, is work on learning affordances for human-robot communication. For example, \citet{kalkan2014verb,ugur2009predicting} learn affordance representations based on the state changes of objects, but do not encode the full trajectory between states. Also related is work in grounded language in text-only models which investigates models ability to reason about objects through space and time \cite{aroca2021prost}.

Outside of NLP, models have been trained on trajectory data for applications like human motion path forecasting \cite{giuliari2021transformer} and human activity recognition \cite{wang2018human}. Our work lies at the intersection of grounded language learning and spatiotemporal machine learning, using representations of trajectory data to study verb semantics.

\paragraph{Grounding and Lexical Semantics.}

Prior work in formal semantics attempts to build feature-based representations of verb meaning in terms of the 3D trajectories and state transitions entailed by those verbs \citep{pustejovsky-krishnaswamy-2014-generating,siskind2001grounding,steedman2002plans}. Such work is related more generally to the idea of mental simulation as a means for representing and reasoning about linguistic concepts \citep{feldman2008molecule,bergen2007spatial,bergen2012louder}. We view our contribution as consistent with and complementary to this formal semantics program. While the prior work has sought to codify the precise truth conditions of motion verbs, we investigate whether such representations could emerge organically from data-driven processes.

While we focus on concrete verbs in this paper, other work has argued that motor processing and mental simulation plays a more general role in language processing. For example, \citet{gardenfors2019using} makes a case for grounded distributional ``conceptual spaces'' as the foundation for modeling linguistic concepts. \citet{dorr2018lexical} discusses the role of metaphor in modeling abstract uses of words like \textit{push}. \citet{borghi2009sentence} argues for the notion of a "motor prototype" as a key component of recognizing and processing objects, and \citet{mazzuca2021affordances} presents evidence that the sensorimotor system (in particular the interactive aspects) drive acquisition of abstract concepts.

\section{Dataset}

\subsection{Overview}

To carry out the proposed study, we require a dataset that contains continuous 3D recordings of an agent interacting with an object. While our representation learning methods will not use linguistic supervision, we require verb labels in order to evaluate our models. Thus, in our data, we require that  each recording is annotated with verbs describing the motion of the object. For example, if the agent throws a bouncy ball across a room, we'd expect the recording to be annotated with a verb sequence such as \textit{be thrown}, \textit{fall}, \textit{bounce}, \textit{bounce}, \textit{bounce}, \textit{roll}, \textit{stop}.

To produce such data, we build a simple Markovian agent which interacts with a variety of objects in a 3D virtual environment. We record the resulting trajectory of the object and then, using crowdsourcing, ask humans to determine which verbs could accurately describe which portions of the object's movement. An example sequence from the dataset is shown in Figure \ref{fig:sequence}.

\subsection{Data Generation and Terminology}\label{sec:datagen}

In this section we provide details on how we generate the data, and introduce terminology that will be used throughout the rest of the paper.

\paragraph{Environment.} The dataset is generated in Unity, a game engine seeing increased use by researchers \cite{deitke2020robothor,gan2020threedworld} for its accessible rendering and physics simulation via the underlying Nvidia PhysX physics engine. The dataset and simulation source code are publicly available.\footnote{\url{https://github.com/dylanebert/simulated}}

\paragraph{Trajectory data.} We define trajectory data as the position and rotation of entities in space, represented with three-dimensional $XYZ$ coordinates and four-dimensional $XYZW$ quaternions respectively. We choose to focus on only these features, ignoring other possibilities like object shape or identity, in order to focus on learning generalizable aspects of verb semantics that are independent of the object.

\paragraph{Sessions.} The dataset is generated in 3-minute continuous segments we refer to as \textbf{sessions}. Within each session, several parameters are randomized, including object shape, mass, drag, friction, and bounciness. 

\begin{figure}
    \centering
    \includegraphics[width=\columnwidth]{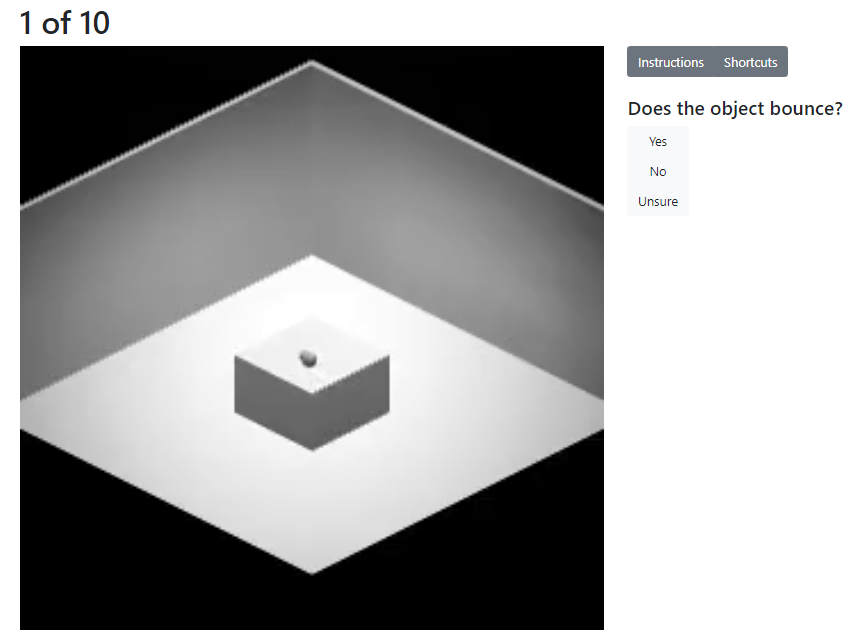}
    \caption{Crowd annotation task. Crowdworkers make binary judgments on whether the verb applies to the clip. In this example, the worker is asked \textit{Does the object bounce?} about the 1.5s video clip on the left.}
    \label{fig:hit}
\end{figure}

\paragraph{Action Primitives.} The data generation is driven by a Markov Chain with a set of randomly parameterized \textbf{action primitives}. In this Markov Chain, the States are whether the object is \textit{Held}, \textit{OnCounter} and \textit{OnGround}. The transitions between these states are action primitives like \textit{PickUp}, \textit{PutDown}, or \textit{Throw}. For example, when the object is in the state \textit{OnCounter}, the agent may execute a \textit{PickUp}, after which the object is \textit{Held}. These action primitives, combined with the physics of the objects (e.g., their shape, mass, friction, bounciness, etc) are intended to produce a wide range of object motions corresponding to a range of verbs, and we do not expect that the primitives will map directly to the verbs that one would use to describe the resulting object behavior. For example, when we simulate a Throw primitive, the result might be that the object flies across the room, hits the wall, falls to the floor, and bounces until it comes to a rest. We parameterize the execution of each action with action-specific parameters, e.g. the force of a throw. The combination of session- and action-level parameters can result in a wide variety of object motion from each primitive action. A full description of the parameters for each action can be found in Appendix \ref{appendix:datasetparams}.

\paragraph{Verbs.} We highlight a distinction between action primitives and the \textit{high-level actions} or \textbf{verbs} that emerge from them. For example, if the object is \textit{pushed}, it may then \textit{slide}, \textit{bounce}, \textit{roll}, \textit{tumble}, or any combination thereof. We refer to all of these as \textit{verbs}, though only \textit{push} is an action primitive. We highlight this distinction because we are most interested in studying the nuanced verbs that emerge from the simulation, rather than the action primitives that drive it explicitly.

\paragraph{Frames.} Our atomic unit is \textbf{frames}, also referred to as \textbf{timesteps}, which represent a single point in time. Our dataset is collected at 60 fps, or 10,800 frames per session. For each frame, we record the position and rotation of the object, as well as the position of the agent. This is sufficient to reconstruct and render the scene from an arbitrary perspective as needed. We choose this high framerate because it's relatively fast and inexpensive to rapidly produce trajectory data, which can be subsampled as needed for rendering or modeling.

\begin{figure*}[ht!]
    \centering
    \includegraphics[width=\textwidth]{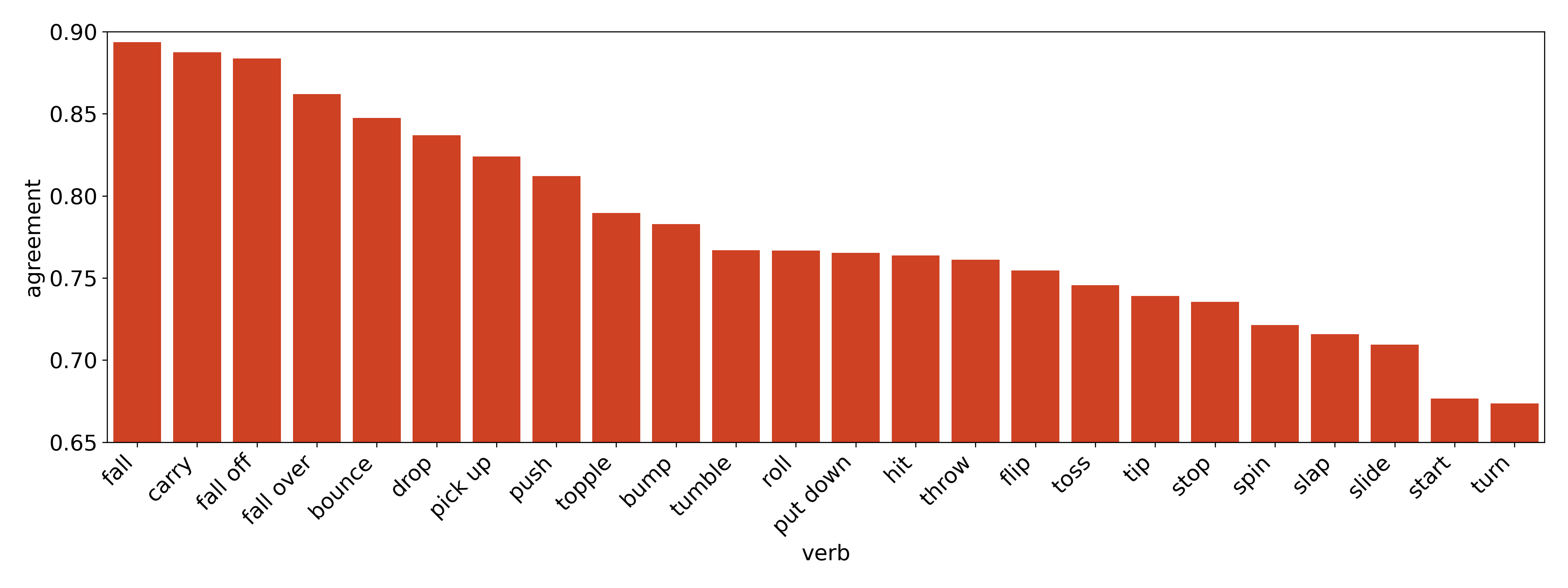}
    \caption{Crowd annotation agreement by verb. Workers agree most on when verbs of gravity occur, such as \textit{fall, drop, bounce}, and least on when verbs of rotation occur, i.e. \textit{turn, spin, tip}.}
    \label{fig:agreement}
\end{figure*}

\subsection{Crowdsourced Annotation}\label{sec:crowdsourcing}

We collect labels for which \textit{verbs} occur in the data, and when they occur. To do this, we extract short clips from the dataset, and ask crowdworkers to provide binary judgments on whether the clip falls in the extension of the verb.

\paragraph{Clips.} We extract short clips from the dataset using Hierarchical Dynamic Clustering with Motion energy-based pooling \cite{zhang2018human}, a self-supervised action segmentation framework that can be summarized as follows:

\begin{enumerate}[noitemsep]
    \item The 3D space is divided into clusters using the provided trajectory data. The framework uses Hierarchical Dynamic Clustering, which is similar to k-means but shown to outperform it on human motion parsing tasks.
    \item A sliding window is applied to the cluster labels for a given positional sequence. The number of transitions between clusters in a window are defined as its \textit{motion energy}. 
    \item The subsequent motion energy curve is smoothed using a Gaussian kernel with a tuned smoothing factor.
    \item The peaks of the motion energy curve are considered motion \textit{segments}, with lengths varying with respect to the width of the peak.
\end{enumerate}

This algorithm is shown to perform well on human motion parsing, which we find transfers well to our dataset when applied to \textit{object position}. This yields easily identifiable patterns of motion, e.g. from the time the object is thrown to when it slows to a stop. We find that, in contrast to a random sliding window, this approach avoids cutting clips in the middle of salient patterns of motion.

In our case, a disadvantage of this approach is that the extracted segments are variable-length. To simplify our pipeline, we filter to only segments of length 72 to 96, then crop the segment to length 90, or 1.5 seconds. We call each 1.5s segment a \textbf{clip}. We choose this length to make the clip as short as possible to avoid crowdworker fatigue, but give sufficient time for a human observer to recognize what's happening.

\paragraph{Verbs.} We produce 24 queries, each corresponding to a verb, e.g. \textit{Does the object bounce?} To do this, the authors curate a list of 24 verbs\footnote{fall, carry, fall off, fall over, bounce, drop, pick up, push, topple, bump, tumble, roll, put down, hit, throw, flip, toss, tip, stop, spin, slap, slide, start, turn} of interest which are likely to occur in the simulated data and range from general descriptions (e.g., \textit{fall}) to more subtle descriptions of object motion (e.g., \textit{tumble}). When asking annotators whether a verb applies to a clip, we always frame the question with the object as the subject. That is, when a carry event occurs, annotators are asked ``is the object carried''.

We then consider every possible (clip, query) pair a potential crowdsourcing \textit{task}. We apply conservative heuristics to filter out (clip, query) pairs that are guaranteed to have a negative label. For example, if the \textit{Held} state was never present in a clip, we don't ask if the object is \textit{carried}. This results in approximately 110k tasks, from which we sample 100 tasks per query, for a total 2400 crowdsourcing tasks, such as the one shown in Figure \ref{fig:hit}.

\paragraph{Labels.} For each crowdsourcing task, we obtain responses from five workers, then take the majority response as the \textit{label} for that clip. The same clip is shown for all applicable queries, resulting in a supervised dataset of 24-dimensional vectors, representing binary verb labels for each clip.\footnote{Labels are only yes or no. \textit{Unsure} was not the majority label for any task. Tasks that were filtered out during crowdsourcing are assigned a mask value that is ignored during training and validation.} The dataset and all unaggregated annotations are available for download.\footnote{\url{https://lunar.cs.brown.edu/simulated/}}

\begin{figure*}
    \centering
    \includegraphics[width=0.7\textwidth]{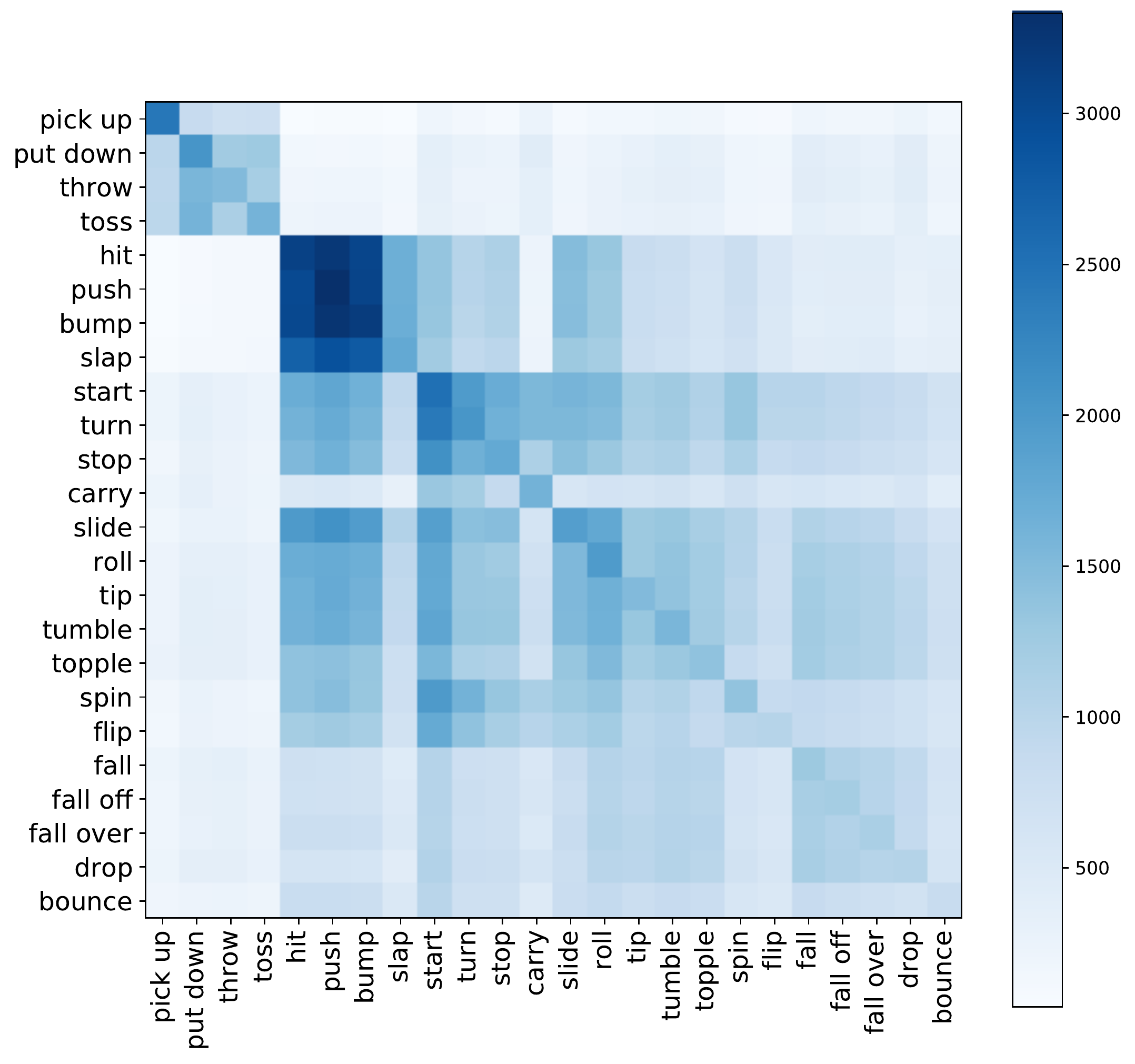}
    \caption{Co-occurrence of different verbs for the same clip. Specifically, given that at least one worker labeled a clip as $v_1$ (row), how many times did other workers label the clip as $v_2$ (column).}
    \label{fig:heatmap}
\end{figure*}

\section{Dataset Analysis}

In this section, we analyze trends in the dataset annotations, including worker agreement, and comparisons between semantically related verbs.

\subsection{Agreement}

Annotation agreement on a clip is the proportion of responses that match the majority label for that clip. Figure \ref{fig:agreement} shows annotation agreement by verb. A noticeable trend is that agreement is higher for particular semantic categories. Specifically, verbs that involve gravity, i.e. \textit{fall}, \textit{fall off}, \textit{drop}, and \textit{bounce} have higher agreement. On the other hand, verbs of rotation, i.e. \textit{turn}, \textit{spin}, \textit{tip}, \textit{flip} have lower agreement, alongside abstract verbs \textit{start} and \textit{stop}. For \textit{start} in particular, we even received feedback from crowdworkers that they weren't sure whether the object \textit{started} moving during the clip or not.

\subsection{Co-occurrence}

Figure \ref{fig:heatmap} shows co-occurrence: specifically, given that a clip is labeled by at least one worker as verb $v_1$, how often is it labeled by other workers as verb $v_2$? Co-occurrence allows us to answer questions like \textit{how often is a \textit{toss} considered a \textit{throw}?} and vice-versa. We highlight some interesting verb relationships.

\paragraph{General co-occurrence.} Verb co-occurrence is high in general. The average number of verbs used to describe a given clip is 4 (where a verb is considered ``used'' if at least three workers use it). This highlights the challenge of verb learning, as opposed to more concrete nouns and adjectives. Verbs are applicable to a wide variety of behavior, even if it isn't a prototypical instance of that verb.

\paragraph{Lexical entailments.} All dogs are animals but not all animals are dogs. These types of semantic containment relationships are also ascribed to verbs. Analyzing our collected data, in some cases, we observe the opposite of what's expected. For example, according to WordNet \cite{fellbaum2010wordnet}, \textit{toss} is a type of \textit{throw}. However, using the majority labels, we find \textit{throws} to be annotated as \textit{tosses} more often \textit{tosses} than are annotated as \textit{throws}. That is, $p(toss|throw)=.67<p(throw|toss)=.75$.

\paragraph{Frequent co-occurrences.} \textit{Hit}, \textit{push}, and \textit{bump} stand out as the most frequently co-occurring verbs, having over 90\% co-occurrence with each other. These likewise occur when many other verbs do, but not reciprocally. For example, most \textit{slaps} are \textit{hits}, but only 41\% of \textit{hits} are \textit{slaps}. In many cases, this can be explained by other verbs being immediately preceded by the agent making contact with the object, which gets labeled \textit{hit}, \textit{push}, and \textit{bump}.

\paragraph{Fine-grained distinctions.} Workers distinguish \textbf{roll} from \textbf{slide} - only 50\% of \textit{rolls} are also considered \textit{slides}, and vice-versa. This validates that verbs with similar trajectories, which may be challenging for models, are indeed differentiated by humans. Additionally, verbs with similar but nuanced meanings are differentiated. For example, \textit{tip}, \textit{tumble}, \textit{fall over}, and \textit{topple} tend to co-occur around ~70-80\% of the time. These also fall into ``verbs of rotation" category, which have the lowest annotator agreement. It isn't clear the extent to which these are nuanced distinctions, or annotation noise.




\begin{figure*}
    \centering
    \includegraphics[width=\textwidth]{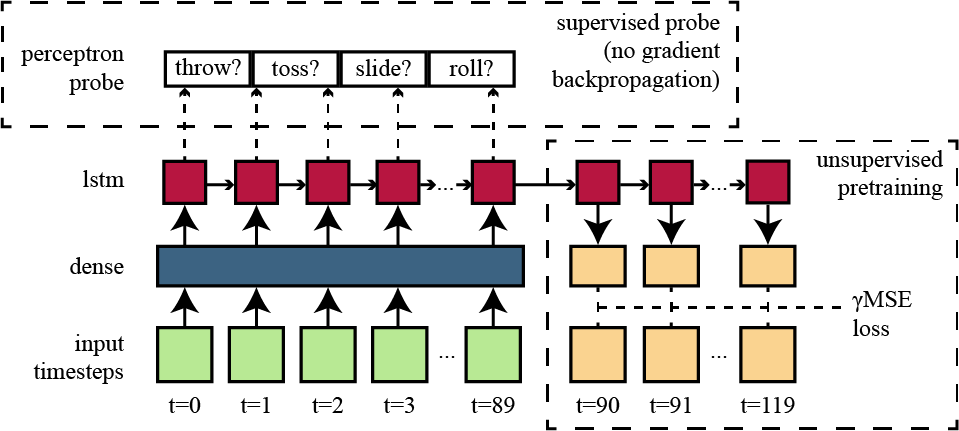}
    \caption{Our pretraining setup. During pretraining, the model learns to encode and represent input timesteps for time-series prediction. To evaluate these learned representations, a perceptron probe is trained on the LSTM outputs, without propagating gradients to the pretrained model.}
    \label{fig:lstm}
\end{figure*}

\section{Experiments}

Our hypothesis is that representation learning in the 3D visuospatial world (without language supervision) can yield concept representations that align to English verb semantics--i.e. can the representations capture nuanced distinctions like \textit{throw} vs. \textit{toss} or \textit{slide} vs. \textit{roll}? To test this, we pretrain a self-supervised model on a time-series prediction task, and then use a perceptron classifier to evaluate its learned representations. 

We evaluate four approaches, described in detail below. First, we train a simple perceptron to evaluate the representational capacity of the trajectory data as-is, as a comparative baseline. Second, we train a fully supervised model to determine a soft upper bound on the task without pretraining. Third, we evaluate our self-supervised model. And finally, we fine-tune the self-supervised model to determine an upper bound with pretraining. 

\subsection{Experimental Setup}

For all approaches, we evaluate representation quality with a multi-way verb classification task. Specifically, we predict the verb labels for the 1.5s clips gathered through the crowdsourcing task described in Section \ref{sec:crowdsourcing}.

Each input sample $X_{t_{1..90}}$ is a 90x10 matrix of position and rotation data, corresponding to 90 frames per clip and 10 spatial features\footnote{Object Position XYZ, Hand Position XYZ, and Object Rotation XYZW.} per frame. The output $Y$ is a 24-dimensional multi-hot vector indicating the whether each of our 24 verb classes apply to the clip.

\subsection{Approaches}

\paragraph{Perceptron.} We wish to evaluate the representational capacity of the raw trajectory data itself. To do so, we train a single 24-dimensional dense layer with sigmoid activation, equivalent to a perceptron for each class. While very simple, this approach gives an idea of how well trajectory data represents verbs as-is, and provides a naive comparative baseline against which to evaluate our more complex pretraining techniques.

\paragraph{Fully Supervised.} The fully supervised approach is similar to the perceptron, but adds a dense layer and LSTM layer in-between. This is equivalent to the model shown in Figure \ref{fig:lstm}, but trained end-to-end without pretraining. The purpose of this approach is to provide an upper bound to the experimental setup \textit{without} pretraining.

\begin{figure*}
    \centering
    \includegraphics[width=\textwidth]{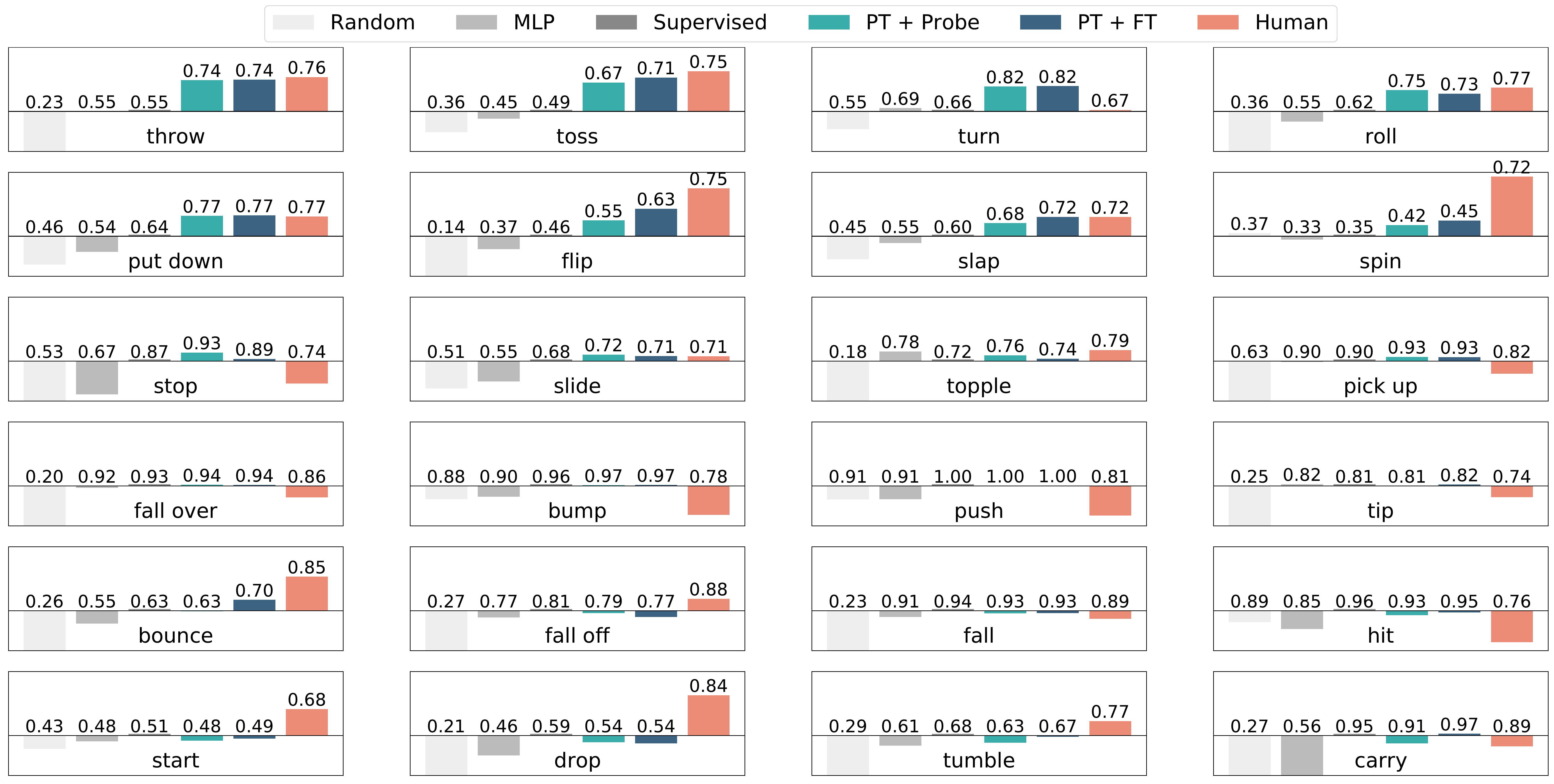}
    \caption{Comparison of each approach by verb. As we are primarily interested in the effect of pretraining, all bars are shown relative to the performance of the fully supervised model. I.e., the fully supervised model (which represents a soft upper bound on how well one can do given raw state information without any pretraining) is visualized as 0, and all other models are visualized relative to that. Human performance is inner-annotator agreement (\% of workers who agree on the majority label).}
    \label{fig:results}
\end{figure*}

\paragraph{Self-supervised Pretraining.} To evaluate the capacity of self-supervised models to represent trajectory data, we pretrain a time-series prediction model on a large unlabeled dataset of 400k sessions. That is, given $n$ input frames $X_{t_{1..n}}$, the model is trained to predict $k$ output frames $Y_{t_{n+1..n+k}}$. The model consists of a dense layer followed by an LSTM layer unrolled $k$ timesteps, as shown in Figure \ref{fig:lstm}. We use a discounted mean squared error loss as shown in Equation \ref{eq:naiveloss}, which discounts loss by how far it is into the future by factor $\gamma$. 

\begin{equation}
    \gamma\text{MSE} = \sum_{t=n}^{n+k} \gamma^t(y_t-\hat{y}_t)^2
    \label{eq:naiveloss}
\end{equation}

\noindent We tune discount factor $\gamma$, output length $k$, model width, and batch size using a grid search on validation performance, resulting in values of 0.85, 60, 128, and 1024, respectively. Input length $n$ is fixed at 90 to match the length of clips.

We consider the concatenated LSTM outputs as the \textit{representation} of a clip. To evaluate this representation compared to raw trajectory data, we freeze the weights of the pretrained model and, as when evaluating the raw trajectory data, train a perceptron for each class.

\paragraph{Fine-tuning.} To provide an upper bound for our experimental setup \textit{with} pretraining, we fine-tune the self-supervised model. This is the same as the previous approach, but allows the gradients in the perceptron step to pass through the entire model.

\section{Results}

\begin{table}[]
    \centering
    \begin{tabular}{l|c}
        \hline
        Approach & mAP (\%) \\
        \hline
        Random Stratified & 41.4 \\
        Perceptron & 65.3 \\
        Fully Supervised & 72.2 \\
        Pretraining + Probe & 76.3 \\
        Pretraining + Finetuning & 77.4 \\
        \hline
    \end{tabular}
    \caption{Mean Average Precision (mAP) for each approach. The pretrained approaches outperform others on verb classification.}
    \label{tab:results}
\end{table}

We report Mean Average Precision on unseen test data for each approach in Table \ref{tab:results}. We compare these to random stratified predictions that are based on the class distribution of the training data. 

\paragraph{Perceptron.} The perceptron approach evaluates the representational capacity of raw trajectory data as-is, with a lower bound of random stratified and soft upper bound of fully supervised. The perceptron performs relatively well for its simplicity, being only 7 points below the fully supervised upper bound. This suggests that the trajectory data itself encodes a significant amount of verb meaning, but leaves plenty of room for improvement.

\paragraph{Self-supervised pretraining.} The pretraining + probe approach evaluates the ability of self-supervised models to encode verb meaning from trajectory data. This is equivalent to the perceptron approach, but with learned hidden representations as input rather than raw trajectory data. The pretrained model does outperform the perceptron, as well as the fully supervised approach. Fine-tuning only improves on this slightly, highlighting that self-supervised pretraining can yield representations that successfully encode verb meaning.

\paragraph{Breakdown by verb.} Figure \ref{fig:results} shows a comparison of average precision for each verb. There are some patterns worth highlighting. In particular, we can categorize verbs into three main groups: \textit{trivial}, \textit{tractable}, and \textit{hard}.

\textit{Trivial} verbs are verbs that can are well-represented by trajectory data as-is, i.e. those with high performance with the perceptron approach. These include \textit{fall, fall off, fall over} and \textit{pick up}.\footnote{We exclude \textit{hit, push}, and \textit{bump} from trivial verbs, as these have high average precision with random stratified, showing that they are very positively skewed, not necessarily well-represented.}. Many of these have high agreement, and may be explained by the object's change in height.

\textit{Tractable} verbs are those that see significant benefit from pretraining, including \textit{slide, roll, throw, toss, put down, turn, flip}, and \textit{stop}. An intuition behind this is that these verbs involve manner distinctions, and in particular, rotations of the object relative to itself. Such information doesn't fall directly out of raw state descriptions, but is likely to be well modeled by a pretraining objective that tries to predict the object's future position.

\textit{Hard} verbs are those with low performance that don't benefit much from pretraining. These include \textit{bounce, drop, tip, topple}, and \textit{spin}. Many of these are verbs which have lower agreement. \textit{Bounce, slap} and \textit{spin} appear to benefit a bit from both pretraining and fine-tuning, suggesting that they may be tractable with similar but more robust pretraining. \textit{Tip} and \textit{topple} have fairly high performance, and may almost be categorized as trivial, perhaps being explained by the object's change in rotation. However, they are noticeably lower than other trivial verbs, despite seeing no benefit from pretraining, suggesting that there is nuance to their meaning in the dataset, which isn't captured by any approach. Finally, \textit{drop} is a great example of a hard verb, as it is similar to trivial verbs like \textit{fall}. However, \textit{drop} involves interaction between the agent and object that is highly agreed upon by annotators, but doesn't appear to be captured by our approaches, despite the model receiving both object and agent data. More challenging examples may be able to unveil a similar story for other verbs of interaction like \textit{pick up} and \textit{put down}.

\section{Discussion and Conclusion}

We test the hypothesis that verb meanings can be grounded in 3D trajectories, i.e., the position of objects over time. Specifically, we investigate the extent to which representations of object trajectories, learned without any linguistic supervision, naturally encode concepts that align to English verb semantics. Our primary contributions are twofold. First, we build a procedurally
generated agent-object-interaction dataset for which we collect crowdsourced annotations. This is the first dataset of its kind, and provides a rich inventory of human judgments about the extensions of 24 verbs of motion. Second, we compare a variety of representation learning approaches, specifically contrasting approaches which operate directly on perceptual inputs to approaches which learn abstractions over the raw perception (via pretraining). We find that some verbs meanings (e.g., \textit{fall} and \textit{push}) are captured easily by the raw state information, while others (e.g., \textit{roll} and \textit{turn}) require additional processing to be represented well.

This work is a first step toward exploring ways to capture fine-grained distinctions in grounded verb semantics that are trivial for humans, but challenging for models. Recent benchmarks at the intersection of NLP, vision and robotics \cite{deitke2020robothor, shridhar2020alfred} illuminate unsolved challenges in AI that demand a more robust understanding of verb semantics and spatial reasoning. As these benchmarks continue to be developed, and rich multimodal datasets from technologies like virtual reality become increasingly abundant, we envision that future work in this vein will be especially relevant.

In the future, we plan to explore more sophisticated models for self-supervised pretraining, and evaluate how well these models transfer to more naturalistic language learning settings \citep{ebert-pavlick-2020-visuospatial}. Beyond this, there is a large body of related research questions to be explored. For example, can representations of trajectory data be fused with visually-grounded representations to yield better encodings of verb semantics? Collaborative efforts will be key to addressing these next milestones in natural language understanding.

\section*{Acknowledgments}

This work was supported by the DARPA GAILA program. Thank you to George Konidaris, Carsten Eickhoff, Roman Feiman, Jack Merullo, Charles Lovering, Gabor Brody, and the members of the Brown LUNAR lab for helpful discussion and feedback. 

\bibliographystyle{acl_natbib}
\bibliography{anthology,acl}

\begin{thebibliography}{37}
\expandafter\ifx\csname natexlab\endcsname\relax\def\natexlab#1{#1}\fi

\bibitem[{Aroca-Ouellette et~al.(2021)Aroca-Ouellette, Paik, Roncone, and
  Kann}]{aroca2021prost}
St{\'e}phane Aroca-Ouellette, Cory Paik, Alessandro Roncone, and Katharina
  Kann. 2021.
\newblock Prost: Physical reasoning of objects through space and time.
\newblock \emph{arXiv preprint arXiv:2106.03634}.

\bibitem[{Bender and Koller(2020)}]{bender-koller-2020-climbing}
Emily~M. Bender and Alexander Koller. 2020.
\newblock \href {https://doi.org/10.18653/v1/2020.acl-main.463} {Climbing
  towards {NLU}: {On} meaning, form, and understanding in the age of data}.
\newblock In \emph{Proceedings of the 58th Annual Meeting of the Association
  for Computational Linguistics}, pages 5185--5198, Online. Association for
  Computational Linguistics.

\bibitem[{Bergen(2012)}]{bergen2012louder}
Benjamin~K Bergen. 2012.
\newblock \emph{Louder than words: The new science of how the mind makes
  meaning}.
\newblock Basic Books.

\bibitem[{Bergen et~al.(2007)Bergen, Lindsay, Matlock, and
  Narayanan}]{bergen2007spatial}
Benjamin~K Bergen, Shane Lindsay, Teenie Matlock, and Srini Narayanan. 2007.
\newblock Spatial and linguistic aspects of visual imagery in sentence
  comprehension.
\newblock \emph{Cognitive science}, 31(5):733--764.

\bibitem[{Bisk et~al.(2020)Bisk, Holtzman, Thomason, Andreas, Bengio, Chai,
  Lapata, Lazaridou, May, Nisnevich, Pinto, and
  Turian}]{bisk-etal-2020-experience}
Yonatan Bisk, Ari Holtzman, Jesse Thomason, Jacob Andreas, Yoshua Bengio, Joyce
  Chai, Mirella Lapata, Angeliki Lazaridou, Jonathan May, Aleksandr Nisnevich,
  Nicolas Pinto, and Joseph Turian. 2020.
\newblock \href {https://doi.org/10.18653/v1/2020.emnlp-main.703} {Experience
  grounds language}.
\newblock In \emph{Proceedings of the 2020 Conference on Empirical Methods in
  Natural Language Processing (EMNLP)}, pages 8718--8735, Online. Association
  for Computational Linguistics.

\bibitem[{Borghi and Riggio(2009)}]{borghi2009sentence}
Anna~M Borghi and Lucia Riggio. 2009.
\newblock Sentence comprehension and simulation of object temporary, canonical
  and stable affordances.
\newblock \emph{Brain Research}, 1253:117--128.

\bibitem[{Brown et~al.(2020)Brown, Mann, Ryder, Subbiah, Kaplan, Dhariwal,
  Neelakantan, Shyam, Sastry, Askell, Agarwal, Herbert{-}Voss, Krueger,
  Henighan, Child, Ramesh, Ziegler, Wu, Winter, Hesse, Chen, Sigler, Litwin,
  Gray, Chess, Clark, Berner, McCandlish, Radford, Sutskever, and
  Amodei}]{DBLP:journals/corr/abs-2005-14165}
Tom~B. Brown, Benjamin Mann, Nick Ryder, Melanie Subbiah, Jared Kaplan,
  Prafulla Dhariwal, Arvind Neelakantan, Pranav Shyam, Girish Sastry, Amanda
  Askell, Sandhini Agarwal, Ariel Herbert{-}Voss, Gretchen Krueger, Tom
  Henighan, Rewon Child, Aditya Ramesh, Daniel~M. Ziegler, Jeffrey Wu, Clemens
  Winter, Christopher Hesse, Mark Chen, Eric Sigler, Mateusz Litwin, Scott
  Gray, Benjamin Chess, Jack Clark, Christopher Berner, Sam McCandlish, Alec
  Radford, Ilya Sutskever, and Dario Amodei. 2020.
\newblock \href {http://arxiv.org/abs/2005.14165} {Language models are few-shot
  learners}.
\newblock \emph{CoRR}, abs/2005.14165.

\bibitem[{Bruni et~al.(2011)Bruni, Tran, and
  Baroni}]{bruni-etal-2011-distributional}
Elia Bruni, Giang~Binh Tran, and Marco Baroni. 2011.
\newblock \href {https://www.aclweb.org/anthology/W11-2503} {Distributional
  semantics from text and images}.
\newblock In \emph{Proceedings of the {GEMS} 2011 Workshop on {GE}ometrical
  Models of Natural Language Semantics}, pages 22--32, Edinburgh, UK.
  Association for Computational Linguistics.

\bibitem[{Chao et~al.(2018)Chao, Liu, Liu, Zeng, and Deng}]{chao2018learning}
Yu-Wei Chao, Yunfan Liu, Xieyang Liu, Huayi Zeng, and Jia Deng. 2018.
\newblock \href {http://arxiv.org/abs/1702.05448} {Learning to detect
  human-object interactions}.

\bibitem[{Deitke et~al.(2020)Deitke, Han, Herrasti, Kembhavi, Kolve, Mottaghi,
  Salvador, Schwenk, VanderBilt, Wallingford et~al.}]{deitke2020robothor}
Matt Deitke, Winson Han, Alvaro Herrasti, Aniruddha Kembhavi, Eric Kolve,
  Roozbeh Mottaghi, Jordi Salvador, Dustin Schwenk, Eli VanderBilt, Matthew
  Wallingford, et~al. 2020.
\newblock Robothor: An open simulation-to-real embodied ai platform.
\newblock In \emph{Proceedings of the IEEE/CVF Conference on Computer Vision
  and Pattern Recognition}, pages 3164--3174.

\bibitem[{Devlin et~al.(2019)Devlin, Chang, Lee, and
  Toutanova}]{devlin-etal-2019-bert}
Jacob Devlin, Ming-Wei Chang, Kenton Lee, and Kristina Toutanova. 2019.
\newblock \href {https://doi.org/10.18653/v1/N19-1423} {{BERT}: Pre-training of
  deep bidirectional transformers for language understanding}.
\newblock In \emph{Proceedings of the 2019 Conference of the North {A}merican
  Chapter of the Association for Computational Linguistics: Human Language
  Technologies, Volume 1 (Long and Short Papers)}, pages 4171--4186,
  Minneapolis, Minnesota. Association for Computational Linguistics.

\bibitem[{Dorr and Olsen(2018)}]{dorr2018lexical}
Bonnie Dorr and Mari~Broman Olsen. 2018.
\newblock Lexical conceptual structure of literal and metaphorical spatial
  language: A case study of “push”.
\newblock In \emph{Proceedings of the First International Workshop on Spatial
  Language Understanding}, pages 31--40.

\bibitem[{Ebert and Pavlick(2020)}]{ebert-pavlick-2020-visuospatial}
Dylan Ebert and Ellie Pavlick. 2020.
\newblock \href {https://www.aclweb.org/anthology/2020.starsem-1.16} {A
  visuospatial dataset for naturalistic verb learning}.
\newblock In \emph{Proceedings of the Ninth Joint Conference on Lexical and
  Computational Semantics}, pages 143--153, Barcelona, Spain (Online).
  Association for Computational Linguistics.

\bibitem[{Feldman(2008)}]{feldman2008molecule}
Jerome Feldman. 2008.
\newblock \emph{From molecule to metaphor: A neural theory of language}.
\newblock MIT press.

\bibitem[{Fellbaum(2010)}]{fellbaum2010wordnet}
Christiane Fellbaum. 2010.
\newblock Wordnet.
\newblock In \emph{Theory and applications of ontology: computer applications},
  pages 231--243. Springer.

\bibitem[{Gan et~al.(2020)Gan, Schwartz, Alter, Schrimpf, Traer, De~Freitas,
  Kubilius, Bhandwaldar, Haber, Sano et~al.}]{gan2020threedworld}
Chuang Gan, Jeremy Schwartz, Seth Alter, Martin Schrimpf, James Traer, Julian
  De~Freitas, Jonas Kubilius, Abhishek Bhandwaldar, Nick Haber, Megumi Sano,
  et~al. 2020.
\newblock Threedworld: A platform for interactive multi-modal physical
  simulation.
\newblock \emph{arXiv preprint arXiv:2007.04954}.

\bibitem[{G{\"a}rdenfors(2019)}]{gardenfors2019using}
Peter G{\"a}rdenfors. 2019.
\newblock Using event representations to generate robot semantics.
\newblock \emph{ACM Transactions on Human-Robot Interaction (THRI)},
  8(4):1--21.

\bibitem[{Giuliari et~al.(2021)Giuliari, Hasan, Cristani, and
  Galasso}]{giuliari2021transformer}
Francesco Giuliari, Irtiza Hasan, Marco Cristani, and Fabio Galasso. 2021.
\newblock Transformer networks for trajectory forecasting.
\newblock In \emph{2020 25th International Conference on Pattern Recognition
  (ICPR)}, pages 10335--10342. IEEE.

\bibitem[{Ji et~al.(2019)Ji, Krishna, Fei-Fei, and Niebles}]{ji2019action}
Jingwei Ji, Ranjay Krishna, Li~Fei-Fei, and Juan~Carlos Niebles. 2019.
\newblock \href {http://arxiv.org/abs/1912.06992} {Action genome: Actions as
  composition of spatio-temporal scene graphs}.

\bibitem[{Kalkan et~al.(2014)Kalkan, Dag, Y{\"u}r{\"u}ten, Borghi, and
  {\c{S}}ahin}]{kalkan2014verb}
Sinan Kalkan, Nilg{\"u}n Dag, Onur Y{\"u}r{\"u}ten, Anna~M Borghi, and Erol
  {\c{S}}ahin. 2014.
\newblock Verb concepts from affordances.
\newblock \emph{Interaction Studies}, 15(1):1--37.

\bibitem[{Kiela et~al.(2017)Kiela, Conneau, Jabri, and
  Nickel}]{kiela2017learning}
Douwe Kiela, Alexis Conneau, Allan Jabri, and Maximilian Nickel. 2017.
\newblock Learning visually grounded sentence representations.
\newblock \emph{arXiv preprint arXiv:1707.06320}.

\bibitem[{Lazaridou et~al.(2015)Lazaridou, Pham, and
  Baroni}]{lazaridou2015combining}
Angeliki Lazaridou, Nghia~The Pham, and Marco Baroni. 2015.
\newblock \href {http://www.aclweb.org/anthology/N15-1016} {Combining language
  and vision with a multimodal skip-gram model}.
\newblock pages 153--163.

\bibitem[{Mazzuca et~al.(2021)Mazzuca, Fini, Michalland, Falcinelli, Da~Rold,
  Tummolini, and Borghi}]{mazzuca2021affordances}
Claudia Mazzuca, Chiara Fini, Arthur~Henri Michalland, Ilenia Falcinelli,
  Federico Da~Rold, Luca Tummolini, and Anna~M Borghi. 2021.
\newblock From affordances to abstract words: The flexibility of sensorimotor
  grounding.
\newblock \emph{Brain Sciences}, 11(10):1304.

\bibitem[{Pustejovsky and
  Krishnaswamy(2014)}]{pustejovsky-krishnaswamy-2014-generating}
James Pustejovsky and Nikhil Krishnaswamy. 2014.
\newblock \href {https://doi.org/10.3115/v1/S14-1014} {Generating simulations
  of motion events from verbal descriptions}.
\newblock In \emph{Proceedings of the Third Joint Conference on Lexical and
  Computational Semantics (*{SEM} 2014)}, pages 99--109, Dublin, Ireland.
  Association for Computational Linguistics and Dublin City University.

\bibitem[{Radford(2020)}]{radford-2020-seeing}
Benjamin Radford. 2020.
\newblock \href {https://www.aclweb.org/anthology/2020.aespen-1.7} {Seeing the
  forest and the trees: Detection and cross-document coreference resolution of
  militarized interstate disputes}.
\newblock In \emph{Proceedings of the Workshop on Automated Extraction of
  Socio-political Events from News 2020}, pages 35--41, Marseille, France.
  European Language Resources Association (ELRA).

\bibitem[{Regneri et~al.(2013)Regneri, Rohrbach, Wetzel, Thater, Schiele, and
  Pinkal}]{regneri2013grounding}
Michaela Regneri, Marcus Rohrbach, Dominikus Wetzel, Stefan Thater, Bernt
  Schiele, and Manfred Pinkal. 2013.
\newblock Grounding action descriptions in videos.
\newblock \emph{Transactions of the Association for Computational Linguistics},
  1:25--36.

\bibitem[{Shridhar et~al.(2020{\natexlab{a}})Shridhar, Jain, Agarwal, and
  Kleyko}]{shridhar-etal-2020-end}
Kumar Shridhar, Harshil Jain, Akshat Agarwal, and Denis Kleyko.
  2020{\natexlab{a}}.
\newblock \href {https://doi.org/10.18653/v1/2020.sustainlp-1.4} {End to end
  binarized neural networks for text classification}.
\newblock In \emph{Proceedings of SustaiNLP: Workshop on Simple and Efficient
  Natural Language Processing}, pages 29--34, Online. Association for
  Computational Linguistics.

\bibitem[{Shridhar et~al.(2020{\natexlab{b}})Shridhar, Thomason, Gordon, Bisk,
  Han, Mottaghi, Zettlemoyer, and Fox}]{shridhar2020alfred}
Mohit Shridhar, Jesse Thomason, Daniel Gordon, Yonatan Bisk, Winson Han,
  Roozbeh Mottaghi, Luke Zettlemoyer, and Dieter Fox. 2020{\natexlab{b}}.
\newblock Alfred: A benchmark for interpreting grounded instructions for
  everyday tasks.
\newblock In \emph{Proceedings of the IEEE/CVF conference on computer vision
  and pattern recognition}, pages 10740--10749.

\bibitem[{Silberer and Lapata(2012)}]{silberer-lapata:2012:EMNLP-CoNLL}
Carina Silberer and Mirella Lapata. 2012.
\newblock \href {http://www.aclweb.org/anthology/D12-1130} {Grounded models of
  semantic representation}.
\newblock In \emph{Proceedings of the 2012 Joint Conference on Empirical
  Methods in Natural Language Processing and Computational Natural Language
  Learning}, pages 1423--1433, Jeju Island, Korea. Association for
  Computational Linguistics.

\bibitem[{Siskind(2001)}]{siskind2001grounding}
Jeffrey~Mark Siskind. 2001.
\newblock Grounding the lexical semantics of verbs in visual perception using
  force dynamics and event logic.
\newblock \emph{Journal of artificial intelligence research}, 15:31--90.

\bibitem[{Steedman(2002)}]{steedman2002plans}
Mark Steedman. 2002.
\newblock Plans, affordances, and combinatory grammar.
\newblock \emph{Linguistics and Philosophy}, 25(5):723--753.

\bibitem[{Sun et~al.(2019)Sun, Myers, Vondrick, Murphy, and
  Schmid}]{sun2019videobert}
Chen Sun, Austin Myers, Carl Vondrick, Kevin Murphy, and Cordelia Schmid. 2019.
\newblock Videobert: A joint model for video and language representation
  learning.
\newblock \emph{arXiv preprint arXiv:1904.01766}.

\bibitem[{Sun et~al.(2018)Sun, Shrivastava, Vondrick, Murphy, Sukthankar, and
  Schmid}]{sun2018actorcentric}
Chen Sun, Abhinav Shrivastava, Carl Vondrick, Kevin Murphy, Rahul Sukthankar,
  and Cordelia Schmid. 2018.
\newblock \href {http://arxiv.org/abs/1807.10982} {Actor-centric relation
  network}.

\bibitem[{Ugur et~al.(2009)Ugur, Sahin, and Oztop}]{ugur2009predicting}
Emre Ugur, Erol Sahin, and Erhan Oztop. 2009.
\newblock Predicting future object states using learned affordances.
\newblock In \emph{2009 24th International Symposium on Computer and
  Information Sciences}, pages 415--419. IEEE.

\bibitem[{Wang et~al.(2018)Wang, Xu, Cheng, Xia, Yin, and Wu}]{wang2018human}
Lei Wang, Yangyang Xu, Jun Cheng, Haiying Xia, Jianqin Yin, and Jiaji Wu. 2018.
\newblock Human action recognition by learning spatio-temporal features with
  deep neural networks.
\newblock \emph{IEEE access}, 6:17913--17922.

\bibitem[{Yatskar et~al.(2016)Yatskar, Zettlemoyer, and
  Farhadi}]{Yatskar_2016_CVPR}
Mark Yatskar, Luke Zettlemoyer, and Ali Farhadi. 2016.
\newblock Situation recognition: Visual semantic role labeling for image
  understanding.
\newblock In \emph{Proceedings of the IEEE Conference on Computer Vision and
  Pattern Recognition (CVPR)}.

\bibitem[{Zhang et~al.(2018)Zhang, Tang, Sun, and Neumann}]{zhang2018human}
Yan Zhang, Siyu Tang, He~Sun, and Heiko Neumann. 2018.
\newblock Human motion parsing by hierarchical dynamic clustering.
\newblock In \emph{BMVC}, page 269.

\end{thebibliography}

\clearpage

\onecolumn
\appendix

\section{Dataset parameters}\label{appendix:datasetparams}

The following tables describe the session-level and action-level parameters for our procedural data generation protocol described in Section \ref{sec:datagen}.

\begin{table*}[!hb]
    \centering
    \begin{tabular}{|l|l|l|}
        \hline
        Parameter & Description & Random Values \\
        \hline
        Start Location & Initial agent location & \textit{random from waypoint set} \\
        \hline
        Start Rotation & Initial agent rotation in degrees & $(0, 360)$ \\
        \hline
        Target Mesh & Shape of the object & \textit{cube/sphere/capsule/cylinder} \\
        \hline
        Target Position & Initial object location & \textit{random location on counter} \\
        \hline
        Target Rotation & Initial object rotation in degrees & $(0, 360)$ \\
        \hline
        Target Mass & Mass in kg of the object & $(0.1, 10)$ \\
        \hline
        Target Drag & Hinders object motion & $(0, 2)$ \\
        \hline
        Target Angular Drag & Hinders object angular motion & $(0.1, 1)$ \\
        \hline
        Dynamic Friction & Friction when object is moving & $(0, 1)$ \\
        \hline
        Static Friction\footnote{Static friction is limited to only have a magnitude difference of 0.3 from dynamic friction.} & Friction when object is not moving & $(0, 1)$ \\
        \hline
        Bounciness & Energy retained on bounce & $(0, 1)$ \\
        \hline
    \end{tabular}
    \caption{Session-level parameters, which add variety between 3-minute sessions.}
    \label{tab:sessions}
\end{table*}

\begin{table*}[!hb]
    \centering
    \begin{tabular}{|l|l|c|}
        \hline
        Parameter & Description & Values  \\
        \hline
        Pick Speed & Hand velocity for Pick motion & $(1, 3)$ \\
        \hline
        Put Speed & Hand velocity for Put motion & $(1, 3)$ \\
        \hline
        Push Speed & Hand velocity for Push motion & $(1, 3)$ \\
        \hline
        Throw Force & Object force for Throw motion & $(25, 125)$ \\
        \hline
        Hit Force & Object force for Hit motion & $(25, 125)$ \\
        \hline
    \end{tabular}
    \caption{Action-level parameters, which add variety in the execution of each action primitive in the dataset.}
    \label{tab:actions}
\end{table*}

\end{document}